\documentclass[fleqn,10pt]{wlscirep}
\usepackage[utf8]{inputenc}
\usepackage[T1]{fontenc}
\usepackage{multirow}
\usepackage{graphicx}
\usepackage{enumitem}
\title{Mastering Chinese Chess AI(Xiangqi) Without Search}

\author[1]{Juntong Lin}
\author[1]{Zhichao Shu}
\author[1]{Yu Chen\thanks{Corresponding Author: \texttt{alexychen@tencent.com}}}
\affil[1]{LIGHTSPEED STUDIOS, Tencent, China}


\begin{abstract}
We have developed a high-performance Chinese Chess AI that operates without reliance on search algorithms. This AI has demonstrated the capability to compete at a level commensurate with the top 0.1\% of human players. By eliminating the search process typically associated with such systems, this AI achieves a Queries Per Second (QPS) rate that exceeds those of systems based on the Monte Carlo Tree Search (MCTS) algorithm by over a thousandfold and surpasses those based on the AlphaBeta pruning algorithm by more than a hundredfold.
The AI training system consists of two parts: supervised learning and reinforcement learning. Supervised learning provides an initial human-like Chinese chess AI, while reinforcement learning, based on supervised learning, elevates the strength of the entire AI to a new level. Based on this training system, we carried out enough ablation experiments and discovered that 1. The same parameter amount of Transformer architecture has a higher performance than CNN on Chinese chess; 2. Possible moves of both sides as features can greatly improve the training process; 3. Selective opponent pool, compared to pure self-play training, results in a faster improvement curve and a higher strength limit. 4. Value Estimation with Cutoff(VECT) improves the original PPO algorithm training process and we will give the explanation.
\end{abstract}
\begin{document}

\flushbottom
\maketitle
%
%
\thispagestyle{empty}


\section*{Introduction}

Deep reinforcement learning (RL) has seen notable success in developing autonomous agents capable of mastering a variety of tasks, from playing Atari games to solving complex real-world problems\cite{schulman2015trust,schulman2017proximal,ye2020towards,sutton1999reinforcement,mnih2015human,lillicrap2015continuous,mnih2013playing,hasselt2010double,jaderberg2019human},. In particular, its application to board games has led to milestones such as AlphaGo and AlphaZero by DeepMind\cite{silver2016mastering,silver2017mastering}, which defeated human champions in Go, and AlphaStar which achieved grandmaster status in the real-time strategy game StarCraft II\cite{vinyals2019grandmaster}. These accomplishments highlight reinforcement learning's potential to handle intricate strategic decision-making.

Chinese Chess, also known as Xiangqi, is a strategy board game for two players that share similarities with the international chess game. It is one of the most popular board games in China and has been played for centuries, boasting a complexity and strategic depth revered by enthusiast and scholar alike. The standard Xiangqi board is a 10x9 grid, with a distinctive feature—the river—that divides the two opposing camps and impacts piece movement.

Each player begins with a set of 16 pieces: one General (king), two Advisors, two Elephants, two Chariots, two Cannons, two Horses, and five Soldiers. The mentioned pieces each have unique movement rules, such as the Elephant's inability to cross the river and the Cannon's requirement of a 'screen' piece for capture. Victory is achieved by checkmating or stalemating the opponent's General.

Chinese Chess invites a range of dynamic tactics and deeper strategies, with an emphasis on balance between attack and defense, and the meticulous timing of aggression. The game presents an astronomically large state space, estimated at around $10^{50}$ possible game positions. This sheer scale poses considerable challenges for AI, particularly for deterministic search algorithms that are prone to combinatorial explosion.

Traditional Chinese Chess AI approaches have emphasized the use of alpha-beta pruning and other heuristic search methods—effective yet limited by the computational resources necessary to examine extensive game trees. Additionally, the algorithms require handcrafted evaluation functions that encapsulate human understanding of the game, which can be rigid and fail to adapt to novel situations.

Contrasting these search approaches, reinforcement learning (RL) agents—such as those applied in recent successes with AlphaGo and AlphaZero—have demonstrated the potential to learn and adapt from gameplay experience, with minimal human-crafted heuristics. However, applying these methods to Chinese Chess AI confronts two main challenges: one is the efficient encoding of game states in a way that preserves the rich context and enables the neural network to assess board positions effectively; and the other is formulating an RL algorithm that can lean into the expansive strategic depth of the game without being explicit guidance from heuristic searches.

Algorithms based on Monte Carlo Tree Search (MCTS)\cite{coulom2006efficient,kocsis2006bandit,browne2012survey}, such as AlphaGo or KataGo\cite{wu2019accelerating}, typically require numerous simulations to compute the next move. For instance, KataGo simulates 800 times, necessitating at least 800 forward inferences by the neural network. It is well-known that computations involving neural networks are highly resource-intensive. As a result, the application of MCTS-based search algorithms in many real-world scenarios is not practical. Our primary contribution is an attempt to eliminate the use of MCTS, a resource-intensive algorithm, from the production process of board game AI. By optimizing other modules within the algorithmic process, we aim to develop a board game AI that can compete at the level of top human players without relying on MCTS.

This paper addresses these challenges by leveraging the representation capabilities of Transformer architectures and designing a tailored RL algorithm that initially learns from human expert games before refining its policy through self-play. We also innovate in our treatment of opponent moves and the selection of adversarial engagements that drive diverse learning scenarios. The goal is to present an AI capable of discovering deep strategic concepts and subtle tactical motifs in Chinese Chess, ultimately revealing the emergent behaviors that arise from an RL-centric approach.

By delving into this alternative pathway, we hope to expand the horizons for AI advancements in Chinese Chess, providing a model that could, in principle, be extended or modified for other board games and decision-making applications. The next sections will elucidate our methodology, present experimental findings, and explore how these insights can influence the broader AI and machine learning landscapes.


\section*{Methods}
\subsection*{Training Algorithm}
\begin{figure}
    \centering
    \includegraphics[width=\linewidth]{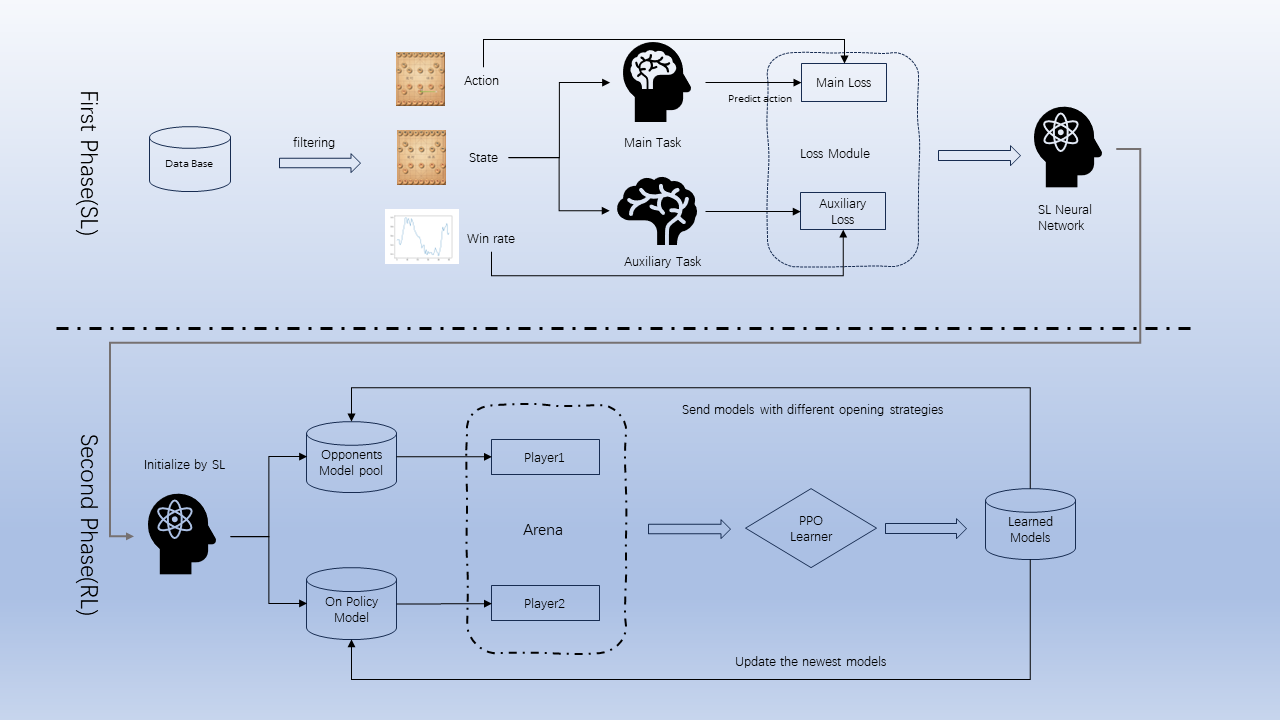}
    \caption{The upper half of the image describes the framework of the training algorithm for supervised learning during the training process, which includes modules such as data cleaning, data sampling, auxiliary tasks, and so on. The lower half of the image describes the algorithmic flow of reinforcement learning during the training process, including restoring the supervised learning model, Dynamics Opponent Pool, and the PPO algorithm, among others.}
    \label{fig:algo_scheme}
\end{figure}

The development of our Chinese Chess AI revolves around a robust training algorithm that addresses the challenges associated with the game's complex action space and strategic depth. The algorithm is divided into two primary phases as shown in Fig.(\ref{fig:algo_scheme}): 1. supervised learning with auxiliary tasks; 2. reinforcement learning using an opponent pool with diverse strategies.

\subsubsection*{Supervised Learning with Auxiliary Tasks}

During the initial phase, we utilize a large dataset of human expert games to instill the AI with a base level of competency in Chinese Chess. Based on meticulously designed sampling curves, we sample from the collected game data to obtain game states that serve as inputs for the neural network. For each game state, there is an associated move selection, which may either be a move chosen by top human players based on the current situation, or a move derived from our calculations using the alpha-beta pruning algorithm. To handle the expansive action space (90x90) more efficiently, we factorize it into two steps. The first is the selection of one of the pieces on the board (up to 16 choices, but fewer if some pieces are captured), and the second is choosing a valid move for that piece (up to 90 choices, but typically fewer based on the game's rules and the current board state). 

\textbf{Sample Curve} 
In the game of Chinese Chess, the distribution of samples is highly unbalanced. Specifically, the entire gameplay process can be divided into the following three stages: the opening, the middle game, and the endgame. The opening refers to the stage from the start of the game to when both sides have essentially formed their initial battle formations. The main objective during this phase is to arrange the pieces logically and develop them rapidly, particularly key offensive and defensive pieces such as chariots, horses, and cannons. During this stage, it is crucial to coordinate and protect the pieces, avoid early mistakes that could lead to the capture of important pieces, establish a solid defense, and look for vulnerabilities in the opponent's defense. The middle game is defined as the stage where both parties have essentially completed their opening setups and begin various tactical maneuvers involving attacks, defenses, and exchanges of pieces. This stage is usually characterized by intense combat and frequent changes in the situation. The endgame refers to the stage where there are fewer pieces on the board, and both sides are contending for the final victory. This phase typically requires high computational power and precise judgment. Therefore, as the game progresses, the demand for computational accuracy increases. Consequently, we have designed the following sampling function: 
$$p \sim \log(a * t + b) + c$$
where $t$ is the game step.

\textbf{Auxiliary Task} We design the auxiliary task to improve the training process. The auxiliary task aims to forecast the win rate after a given move. By predicting the move's outcome, the AI can consider not only immediate tactics but also long-term strategy implications. This is particularly useful given that different pieces have unique movement capabilities and exert varying levels of influence over the game. Hence, the AI learns not only to evaluate specific moves, but also to understand the underlying principles that dictate piece value and positional strength. Besides, a well-learned win rate predictor can be used as the initial value function for the reinforcement learning.

\textbf{Feature With Rule}
Chinese Chess, also known as Xiangqi, is a complex board game that often necessitates extensive computational simulations to ensure that each move made is optimal. This requirement aligns closely with the functionalities of the Monte Carlo Tree Search (MCTS) algorithm. In our research, we have embedded the rules of Chinese Chess into the feature space to enhance the AI's comprehension of the game.

For the input to our neural network, we include not only the current board state but also two additional categories of channels:
1. Based on the current position, identifying which pieces can be moved and to which locations they can be moved.
2. Based on the current position from the opponent's perspective, identifying potential moves and their destinations.

Our subsequent analyses clearly demonstrate that these features significantly expedite the training process of networks based on the ResNet architecture, and they contribute to a notable enhancement in overall performance.

\subsubsection*{Reinforcement Learning}
Following the supervised learning phase, we endeavor to further improve the AI's performance and adaptability by introducing reinforcement learning (RL), which is usually described by Markov Decision Process(MDP). An agent can take some action $a_t$ at time $t$ according to the state $s_t$, and then the environment will transfer to next state $s_{t+1}$ and give a reward $r_t$ to the agent, obeying the dynamics $P(s',r'|s, a)$. The goal of RL is to maximize the accumulated reward along the trajectory, 
$$
\max \sum_{n} \gamma^n r_{t+n}
$$
where $\gamma$ is the discounted factor to guarantee the convergence of the algorithm.
We use the Proximal Policy Optimization (PPO) algorithm\cite{schulman2017proximal}, taking into account its advantages in balancing exploration and exploitation. However, PPO's standard application is modified to suit the deterministic and high-stakes nature of Chinese Chess.
To address the challenges of PPO in the Chinese Chess context, we implement the following adjustments:

\textbf{Opening Strategy Diversification} In the opening phase of Chinese Chess, there are numerous distinct styles of initiating gameplay, ranging from defensive, to offensive, to more balanced approaches. Top-tier AI systems need to adeptly handle these various styles. To prevent the AI from settling into a suboptimal state during self-play, we have designed a concept called "Opening Strategy Diversification." This approach ensures that the AI explores a wide range of strategic possibilities, thereby enhancing its adaptability and robustness in real-game scenarios. Instead of sampling from the policy distribution, we manually curate a set of opening moves from top human players. This diversification in the opening book ensures that our AI is exposed to a variety of game commencements, encouraging a broader strategic understanding from the outset of the game.

\textbf{Dynamic Opponent Pool} It is well recognized that employing self-play as a training method in reinforcement learning can easily lead to cyclic traps, where Model A defeats Model B, Model B defeats Model C, and Model C defeats Model A. Consequently, we have drawn inspiration from the concept of a "League," as utilized in AlphaStar, and have designed a diverse pool of opponent models that serve as opponents during self-play sessions. This pool is continuously updated based on performance metrics and stylistic evaluations, ensuring that our primary AI model is regularly challenged by a wide spectrum of strategic types. The dynamic nature of this pool promotes the emersion of a robust and versatile policy, capable of counteracting diverse strategies.

For each game, one of the opponent models $i$ in the pool is chosen based on the win rate of the current training model against the opponent model $r_{i}$. Strong opponent models (lower win rate) are more likely to be selected. The expression of sampling method is as follows:
$$
p^i_{selected} \varpropto e^{-r_{i} / \tau} 
$$
The opponent model pool is initialized with the recovered model from supervised learning. Current training model will be added to the opponent model pool when its' lowest win rate against models in the pool exceeds a threshold. Moreover, to better challenge the training model, sampling temperature of the opponent model is increased (more like ArgMax) while the temperature of the training model remains unchanged.

\textbf{Value Estimation with Cutoff} 
For PPO training, people usually use GAE(General Adavantage Estimation) to estimate the advantage function\cite{schulman2015high}:
$$
\hat{A}^{GAE}_t = \sum_{n=0}^{+\infty}(\gamma\lambda)^n \delta_{t+n}^V
$$
where $\delta_t^V = r_t + \gamma V(s_{t+1}) - V(s_t)$.

Upon analysis, we identified challenges unique to the use of the PPO algorithm in chess-like adversarial games, particularly concerning the estimation of the value function. Firstly, PPO is an on-policy algorithm that also utilizes entropy loss to balance exploration and exploitation. Consequently, it inevitably samples moves that are exceptionally good or poor during gameplay. This variability can cause the win rate to fluctuate dramatically from 0\% to 100\%, or vice versa, leading to significant volatility in the value function and increasing the neural network's fitting difficulty. Moreover, this phenomenon becomes more pronounced as the game progresses and the number of sampling steps increases. From this observation, it becomes necessary to implement truncation in the trajectory processing, which is called \textbf{Value Estimation with Cutoff}(VECT).


The core algorithm of VECT is:
$$
\hat{A}^{VECT}_t = \sum_{n=0}^{L} (\gamma\lambda)^{n}\delta_{t+n}^V
$$


The above adjustments to the training algorithm are designed to cultivate an AI that not only grasps the objective nuances of Chinese Chess but also achieves a high level of gameplay creativity and adaptability. With these innovative algorithmic enhancements, we aim to elevate the proficiency of our AI to compete at a level indistinguishable from top human players, all in the absence of traditional, computationally demanding search algorithms like MCTS.

\subsection*{Training Data}
In the course of our research, we have amassed a substantial dataset from the LightSpeed and Quantum Studio's "TianTian Xiangqi" platform. It is pertinent to note that during the data collection phase, careful measures were taken to filter out private data such as players' personal information and game details. Only the chess game records were retained, stored in the Portable Game Notation (PGN) format, which adheres to international standards.

The data collection was conducted in two distinct methods. The first method involved gathering data from the top 10 percent of players, encapsulating approximately five million records. For each game scenario, the moves selected by these elite players were specifically documented. The second method expanded the scope to include a broader spectrum of human expertise, ranging from beginners to top-level players, covering ten million game scenarios. In this approach, an agent created using the Alpha-Beta pruning algorithm annotated the moves to enhance the generalizability of the model training.

Furthermore, to ensure the integrity and relevance of the data, the following rules were applied for data cleansing:
1. Elimination of game records where the number of rounds was less than ten.
2. Removal of games terminated due to players disconnecting, which were primarily attributed to network issues.
\subsection*{Model Architecture}

Conceptually, our model maps observation $\mathbf{o}_t$ to action $\mathbf{a}_t$ with a neural network backbone $f_{\mathbf{\theta}}$. 

\textbf{The observation} $\mathbf{o}_t$ is a tensor (of size $H * W * (2 * C)$), including board information (of size $H * W * C$) and (valid move information of size $H * W * C$). $H=10$ and $W=9$ are height and width of the board, $C=7$ denotes number of chess type.For board information, $board_{ij}$ is the one-hot representation of the chess on $i$th row and $j$th column. For valid move information, $validmove_{ijk}$ is a flag indicating whether $i$th row and $j$th column can accessed by chess of type $k$ in one step.

\textbf{The action} $\mathbf{a}_t$ is a one-hot representation of $N=8100$ possible moves (each move can be represented by a pair $<start, end>$, both $start$ and $end$ have at most $90$ choices). Invalid moves are masked in both training and testing.

As for \textbf{the neural network backbones} $f_{\mathbf{\theta}}$, both CNN-based and Transformer-based architectures are verified in this work\cite{krizhevsky2012imagenet, lecun1995convolutional, ioffe2015batch}.

For CNN-based neural netowrk, we adopted the popular ResNet as the backbone\cite{he2016deep}. The observation mentioned above can be viewed as an image and seamlessly fed to ResNet. Considering the scenario of chess, we derive a variant of the original ResNet (denoted by \textbf{mod-ResNet} in the following paper) by modifying the entrance process block. Concretely, conv1 is changed from conv(7x7, 64, 2) to conv(1x1, 64, 1) and the max pool is removed. The insight behind the modification is the importance and necessity of every single "pixel" in the context of chess in comparison with classic image classification task. Experiment results show that the modification brings considerable improvement in intensity.

For Transformer-based, we leverage ViT, which is a transformer-based classification architecture\cite{dosovitskiy2020image}. For the similar motivation in mod-ResNet, we treat each "pixel" of the observation as a token instead of patching neighbouring "pixel" as default. Huge improvement in intensity is achieved in the transition from mod-ResNet to ViT. We attribute the boost to the powerful spatial relation understanding ability of transformer.

\section*{Results}
\subsection*{MainResult}
In this section, we will provide a comprehensive overview of the results obtained from our algorithm. In the main results section, we intend to present the outcomes of our model when pitted against a baseline model, as well as the performance achieved in ranked matches against human opponents. Initially, the evaluation process employed a baseline AI that utilizes the AlphaBeta pruning algorithm. This rule-based AI has demonstrated a level of proficiency approximately within the top 10\% percentile in human ranked competitions. We present five representative milestone results that include their performance against baseline, rankings in human ladder tournaments, and data on predictive accuracy at various stages of the training process, as shown in Table.(\ref{tab:main_result}). Initially, the nomenclature of the five models adheres to a format combining the training algorithm with the model structure. Here, 'SL' denotes Supervised Learning, and 'RL' indicates Reinforcement Learning. The term \textbf{modResNet} refers to our modified version of the ResNet model, adapted specifically for the Chinese Chess scenario, while 'ViT' (Vision Transformer) repurposes the transformer structure commonly utilized in classical visual contexts.

\begin{table}[ht]
\centering
\begin{tabular}{cccccc}
\hline
  \textbf{ModelName} & \textbf{VS baseline} & \textbf{Rank in HumanLadder} & \textbf{Accuracy(First)} & \textbf{Accuracy(Mid)} & \textbf{Accuracy(Last)} \\
  Baseline &  48.5\%(3.12\%) & Top-10\% & NAN & NAN & NAN \\
  SL-modResNet18 & 38.6\%(19.8\%) & NAN & 60.2\% & 47.4\% & 43.3\% \\
  RL-modResNet18 & 52.3\%(9.7\%) & NAN & NAN & NAN & NAN \\
  SL-ViT & 65.1\%(12.8\%) & Top-3\% & 72.6\% & 60.3\% & 59.1\% \\
  RL-ViT & 92.5\%(2.8\%) & Top-0.1\% & NAN & NAN & NAN \\ 
\hline
\end{tabular}
\caption{The first column is category of model, which contains the training algorithm and model architecture. The second column is battle result versus baseline model, in the form of win rate(draw rate). The third column is percentile in the human rank ladder. The last three columns are the accuracy for different stage of game during the training process.}
\label{tab:main_result}
\end{table}

Given the rules of Chinese Chess, which allow for draws at the end of matches, the results against the baseline model are expressed in terms of both win rates and draw rates, with the latter presented within parentheses. It is notable that the sampling function for model matchups is given by:
$$p \sim e^{logits/\tau}$$
where $\tau$ denotes the sampling temperature controlling the entropy of distribution. When battling against the baseline model, we set the temperature at 1 to ensure diversity in evaluation matchups. Our findings indicate that both the training methodology and model architecture optimizations significantly enhance performance. For instance, using reinforcement learning to further optimize the \textbf{modResNet18} architecture increased its win rate against the baseline from 38.6\% to 52.3\%. Similarly, for the ViT structure, reinforcement learning elevated the win rate from 65.1\% under supervised learning to 92.5\%. It is important to note that, due to the evaluation sampling at a temperature of 1, the true win rate of RL-ViT against the baseline is likely much higher than 92.5\%, approaching 100\%.

Beyond confrontations with the baseline model, the ViT-based models, SL-ViT and RL-ViT, were also tested in human ladder tournaments using a sampling temperature of 5 to ensure an accurate measure of their strength. The outcomes were impressive, with SL-ViT achieving a top 3\% ranking and RL-ViT reaching the top 0.1\% level, nearly comparable to top professional players.

Finally, the game was segmented into three phases: First, Mid, and Last. We demonstrated the impact of adjustments in the model structure on overall predictive accuracy within the supervised learning framework. The increase in ResNet size and the adoption of ViT both positively influenced overall accuracy rates.

\subsection*{Ablation Study}
\subsubsection*{Sample Curve}
\begin{figure}
    \centering
    \includegraphics[width=0.5\linewidth]{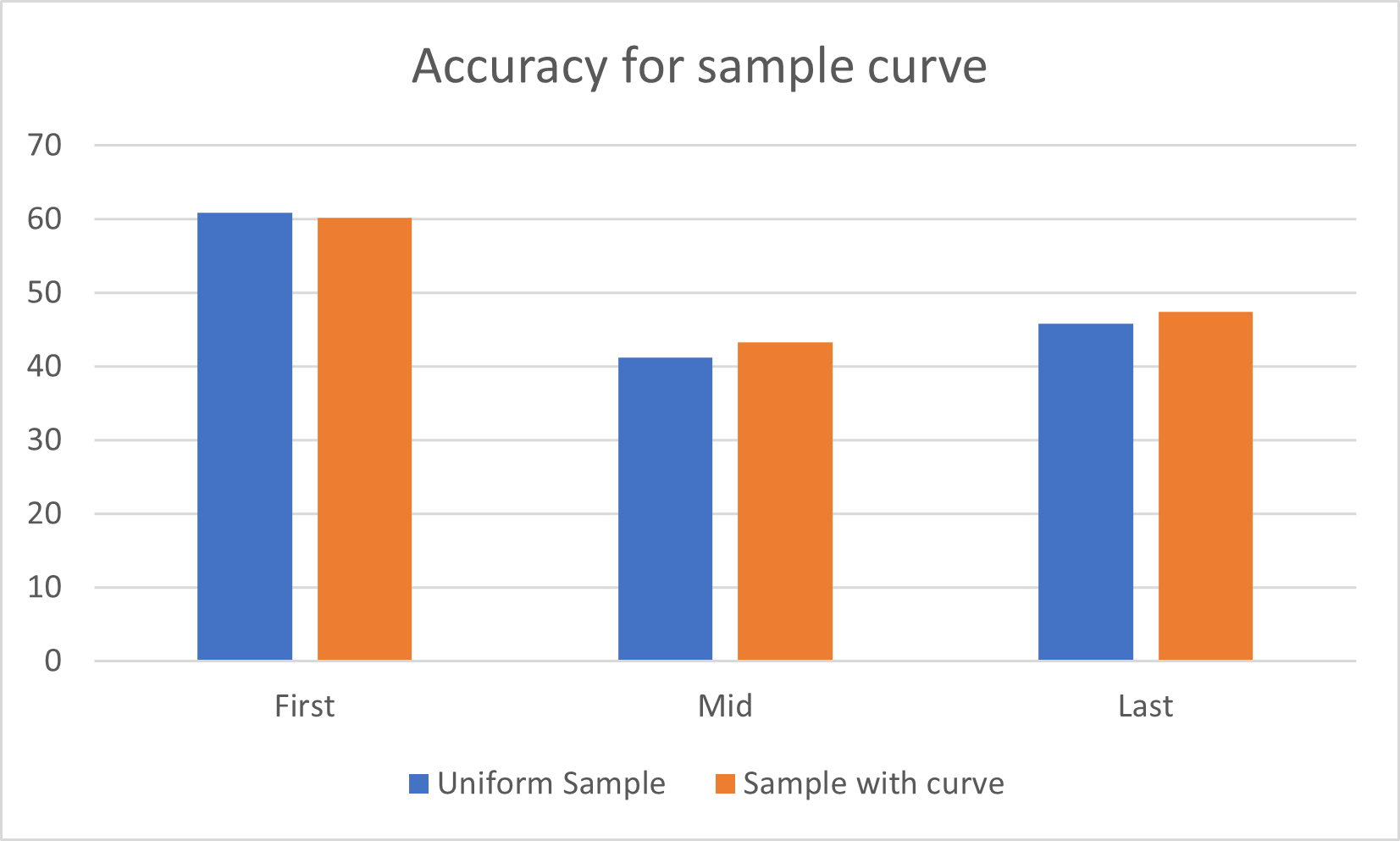}
    \caption{Comparison of accuracy for different game stages between uniform sampling and sample with curve}
    \label{fig:sample_curve}
\end{figure}

The distribution of samples in Chinese Chess is notably imbalanced, and there is a significant disparity in the styles of moves across different phases of the game. Consequently, it is crucial to perform necessary sampling procedures on the training samples. To address this, we designed a specific sampling curve: 
$$p \sim \log(\frac{t}{2T} + e^{-1}) + 1.5$$
where $T$ is the total length of one game varying with each other, $t$ is the current step of the game. This curve illustrates that as the game progresses, the selection of each move requires increasing precision. As demonstrated in Fig.(\ref{fig:sample_curve}), we conducted an ablation study using the \textbf{modResNet18} model, comparing the effects of this sampling curve with uniform sampling during training. We observed that the overall prediction accuracy, particularly in the mid game and last game phases, was enhanced.

\subsubsection*{Feature With Rule}
In this discussion, we explore effective feature modeling for Chinese Chess. The importance of feature modeling may be mitigated if search technologies are employed, as the agent can simulate downwards continuously, ultimately achieving a comprehensive understanding of the current situation. However, in the absence of search and simulation, it necessitates a deeper knowledge of the entire chessboard and the rules of the game on the part of the agent. To this end, we considered the following three methods of feature modeling: 
\begin{itemize}
    \item Features composed solely of the information of the pieces on the board.
    \item Features that include the state of the board and the various move options available to the player.
    \item Features that encompass both the board information and all possible moves for both players based on the current situation.
\end{itemize}

\begin{figure}[htbp]
\centering
\includegraphics[width=0.9\linewidth]{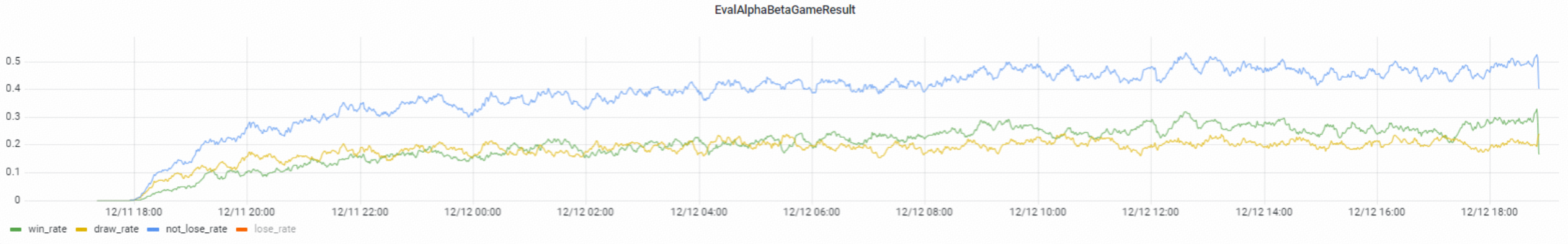}

\vspace{10pt} 

\includegraphics[width=0.9\linewidth]{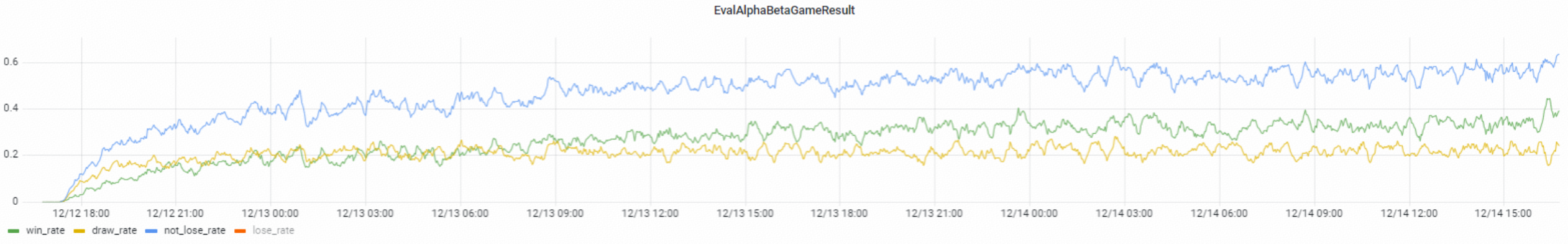}

\vspace{10pt} 

\includegraphics[width=0.9\linewidth]{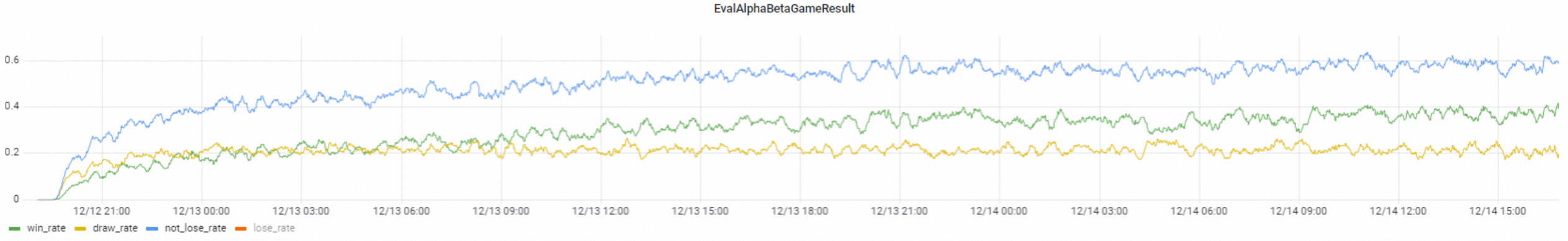}
\caption{Win(Draw) rate for modResNet18 versus baseline with different features. From top to bottom, feature contains game board state, ally valid moves, enemy valid moves step by step}
\label{fig:feature}
\end{figure}

We present the training curves for these three modeling approaches in a graph, where the horizontal axis represents training time, and the vertical axis shows the win and draw rates against a baseline, as shown in Fig.(\ref{fig:feature}). From these training curves, we observe the following interesting phenomena:
\begin{itemize}
    \item The inclusion of permissible moves for the player into the features significantly accelerates the training speed in the early stages. For example, to reach a win rate of 20\%, purely board-based modeling requires 7 hours of training time, whereas incorporating permissible moves reduces this to 5 hours. This aligns closely with the human approach to learning chess: beyond understanding the board, coaching that includes guidance on possible moves can rapidly aid students in discovering more effective strategies.
    \item In the mid to late stages of training, given the same amount of training time, the inclusion of the opponent's permissible moves results in a higher win rate compared to including only the player's own moves. This is because understanding the opponent's countermeasures can help the agent to comprehend the current situation better, ensuring more rational decision-making in the given context.
\end{itemize}
These observations highlight the significant impact of nuanced feature modeling on the performance and efficiency of AI agents in strategic games like Chinese Chess.

\subsubsection*{Model Architecture}
For board games such as Chinese Chess, the state of the board can be naturally represented in a format similar to a 2D image, denoted as $(H, W, C)$, where each channel can represent different types of chess pieces or other information. ResNet, a model originally designed for image processing, can potentially handle such types of encoded information well. However, during our training process, we discovered that directly utilizing ResNet yielded very poor results. Upon deconstructing the ResNet architecture, we identified two processing techniques that are suitable for image scenarios but inappropriate for board game scenarios: 
\begin{itemize}
    \item  A larger kernel size, where a default of (7x7) is typically used.
    \item  Max pooling.
\end{itemize}

These information processing techniques in the context of image handling can be understood as helping the model to capture key features in an image while discarding and merging minor details. Clearly, this approach is not suitable for handling board game information where every piece of information about each chess piece is crucial and does not require merging. Consequently, we modified the traditional ResNet by removing the max pooling and reducing the kernel size to 1x1. The overall experimental results are shown in the Table.(\ref{tab:model_architecture}). From these experimental observations, it is evident that both the performance against the baseline model and the prediction accuracy have significantly improved.

Naturally, derivative structures based on Convolutional Neural Networks (CNNs) often exhibit limitations in processing board game information, specifically in their inability to effectively discern which areas of the board are more critical and which are less so at any given moment. It is well-known that architectures based on Transformers, which rely on underlying attention mechanisms, can adeptly describe the varying degrees of urgency across different regions of the battlefield. In this regard, we employed the Vision Transformer (ViT) network architecture to handle this image-like board information. From the data presented in the Table.(\ref{tab:model_architecture}), it is evident that under the same parameter conditions, ViT significantly enhances overall performance. Notably, there is an approximate 50\% increase in accuracy during the latter stages of the game. Additionally, when competing against the baseline model, the overall win rate improved from 38.6\% to 65.1\%.
\begin{table}[ht]
\centering
\begin{tabular}{ccccc}
\hline
  \textbf{Model Description} & \textbf{VS baseline} &  \textbf{Accuracy(First)} & \textbf{Accuracy(Mid)} & \textbf{Accuracy(Last)} \\
  modResNet18 & 38.6\%(19.8\%) & 60.2\% & 47.4\% & 43.3\% \\
  ResNet18 & 1.9\%(2.1\%) & 57.5\% & 39.8\% & 42.9\% \\
  ResNet18 without pooling & 2.1\%(2.5\%) & 57.4\% & 45.2\% & 43.1\%\\
  ResNet18 with kernel $1\times 1$ & 0.1\%(2.5\%) & 54.1\% & 35.7\% & 38.0\% \\
  ViT & 65.1\%(12.8\%) & 72.6\% & 60.3\% & 59.1\% \\
\hline
\end{tabular}
\caption{ablation study for model architecture}
\label{tab:model_architecture}
\end{table}

\subsubsection*{DOP and VECT}
In this section, we will compare the impact of two predominant methods on the effectiveness of the reinforcement learning training process: DOP and VECT. For adversarial games in general, iterative self-improvement can be achieved through self-play in the reinforcement training process. However, this approach has inherent drawbacks: 1. The opponent model lacks diversity; 2. During the training process, the agent gradually forgets previously learned strategies and methods for countering specific strategies. To mitigate these issues, we employ the DOP method. As evidenced in Table X, the incorporation of DOP has enhanced our overall training outcomes, elevating the win rate against the baseline model from 84\% to 92.5\%.

During the reinforcement learning phase, we initially employed the Generalized Advantage Estimation (GAE) algorithm inherent to the PPO, setting both gamma and lambda to 1. Under these settings, the value function primarily predicts the ultimate win rate of the match, leaning towards using the final outcome as its supervisory signal. However, as indicated in the Table.(\ref{tab:bve}), this configuration does not facilitate effective training in reinforcement learning; the training process tends to crash, with the win rate against the baseline model drastically dropping from 65\% to approximately 10\%. Instead, we use VECT method to estimate the value function, we can find that the win rate is improved from 65\% to 92.5\%, which makes great progress.

\begin{minipage}{0.45\linewidth}
\centering
\begin{tabular}{|c|c|}
\hline
    \textbf{Training Algorithm} &  \textbf{VS baseline} \\
\hline
     Self play &  84.0\%(5.6\%) \\
\hline
     DOP & 92.5\%(2.8\%) \\
\hline
\end{tabular}
\captionof{table}{This table compares the win(draw) rate versus baseline for different training strategy. The first row uses self play, which always select the newest model as the opponent, The second row uses a dynamics opponent pool, which is sampled when choosing opponent model.}
\label{tab:dop}

\end{minipage}
\hfill
\begin{minipage}{0.45\linewidth}
\centering
\begin{tabular}{|c|c|}
\hline
\textbf{Training Algorithm} & \textbf{VS baseline} \\
\hline
GAE & 10.2\%(1.5\%) \\
\hline
VECT & 92.5\%(2.8\%) \\
\hline
\end{tabular}
\captionof{table}{This table compares two value estimation method of RL training for Chinese Chess. The first row uses the classical algorithm GAE. The second row uses a novel method called VECT, which is the improved version of GAE}
\label{tab:bve}
\end{minipage}

\section*{Discussion}

The development and validation of our Chinese Chess AI without search have led to several compelling findings with implications for both reinforcement learning (RL) methods and real-world applications. Here, we discuss the key insights from this research and their potential broader impact. 

Also, we find the work\cite{ruoss2024grandmaster} studying the Chess AI(not Chinese Chess) without search. However, the main contribution of this work focus on the supervised learning and relevant results.

Our results contribute to the growing literature that illuminates the versatility and effectiveness of Transformer architectures. Notably, the performance improvement observed when transitioning from a CNN-based model to a Transformer-based model underscores the superior spatial relation understanding that Transformers offer. This outcome reinforces the idea that attention mechanisms are well-suited to tasks requiring the contextual processing of input features, suggesting that Transformers could have wider applicability in the domain of games with complex strategic elements and perhaps beyond—into areas such as robotics and autonomous systems.

The enhancement of the AI's feature set with rules-based information significantly accelerated and improved the learning process. This underscores the importance of deep domain knowledge when designing input representations for learning models. While this work focused on Chinese Chess, there is potential to extend such feature engineering methods to other domains where expert knowledge can guide the representation of state spaces to aid learning algorithms.

Our method of diversifying opponent strategies in the training phase highlights the limitations of traditional self-play methods and points to the need for more sophisticated training regimens. The concept of using a dynamic pool of opponents may serve as a blueprint for developing RL agents in other adversarial settings, where the strategic depth of the task benefits from exposure to a variety of strategies.

The introduction of the VECT method in the RL phase provides a robust and stable method for evaluating value functions in turn-based game scenarios, particularly in situations where the overall rewards vary significantly. This is extremely beneficial for learning and developing chess AI, as it helps the AI system to more accurately predict the long-term benefits of its actions, thereby enabling it to make better decisions.

Our findings have notable implications for the future of game AI development. By demonstrating the potential to reach high levels of performance without reliance on search algorithms, we pave the way for more lightweight and adaptable AIs in resource-constrained environments. This approach may be especially relevant for developing AIs for real-world applications where computational resources are limited or where low-latency decisions are crucial. The confluence of high performance and reduced computational demands makes for a compelling case for rethinking traditional approaches to game AI development and, by extension, decision-making systems in broader contexts.

Though our results are promising, several limitations need to be addressed in future work. Most notably, our methods were applied in the context of a deterministic and perfect information game—Chinese Chess. As such, the applicability of our techniques to games or scenarios with elements of randomness or imperfect information remains to be demonstrated. Additionally, the AI's performance, while impressive, may still benefit from further refinements in architectural design, training algorithms, and feature engineering.

In conclusion, this research contributes valuable insights into the reinforcement learning landscape and champions methods that could have implications for decision-making systems in general. As we look to future applications and extensions of our work, the blend of RL strategies, model architectures, and feature representations established herein will undoubtedly inform and inspire ongoing developments in AI.

\section*{Acknowledgements}
We thank Tencent Lightspeed studios for opportunity to do research on Chinese Chess. 

\section*{Author Contributions}
Yu Chen and Zhichao Shu set up the whole project for studying Chinese Chess without search. Yu Chen and Juntong Lin designed and implemented the algorithm architecture, evaluation system and wrote the paper. Yu Chen also evaluated the agent in the human ladder rank.

\section*{Author Information}
The authors have disclosed no financial conflicts of interest. Feedback on the online version of the article is encouraged. For correspondence and inquiries regarding materials, please contact alexychen@tencent.com.

\bibliography{sample}

\end{document}